\documentclass[]{spie}  
\usepackage{booktabs}
\usepackage{makecell}

\usepackage{amsmath,amsfonts,amssymb}
\usepackage{multirow}

\usepackage{graphicx}
\usepackage[colorlinks=true, allcolors=blue]{hyperref}
\usepackage{booktabs} 

\title{Evaluation of Intra-operative Patient-specific Methods for Point Cloud Completion for Minimally Invasive Liver Interventions
}

\author[a]{Nakul Poudel}
\author[a]{Zixin Yang}
\author[a]{Kelly Merrell}
\author[b]{Richard Simon}
\author[a,b] {Cristian A. Linte}
\affil[a]{Center for Imaging Science, Rochester Institute of Technology, Rochester, NY 14623, USA}
\affil[b]{Biomedical Engineering, Rochester Institute of Technology, Rochester, NY
14623, USA}

\authorinfo{Further author information: 
\\Nakul Poudel: E-mail: np1140@rit.edu\\
Zixin Yang: E-mail: yy8898@rit.edu\\
Kelly Merrell: E-mail: kam1217@rit.edu\\
Richard Simon: E-mail: rasbme@rit.edu \\
Cristian A. Linte: E-mail: calbme@rit.edu}

\begin{document} 
\maketitle

\begin{abstract}


The registration between the pre-operative model and the intra-operative surface is crucial in image-guided liver surgery, as it facilitates the effective use of pre-operative information during the procedure. However, the intra-operative surface, usually represented as a point cloud, often has limited coverage, especially in laparoscopic surgery, and is prone to holes and noise, posing significant challenges for registration methods. Point cloud completion methods have the potential to alleviate these issues. Thus, we explore six state-of-the-art point cloud completion methods to identify the optimal completion method for liver surgery applications. We focus on a patient-specific approach for liver point cloud completion from a partial liver surface under three cases: canonical pose, non-canonical pose, and canonical pose with noise. The transformer-based method, AdaPoinTr, outperforms all other methods to generate a complete point cloud from the given partial liver point cloud under the canonical pose. On the other hand, our findings reveal substantial
performance degradation of these methods under non-canonical poses and noisy settings, highlighting the limitations of these methods, which suggests the need for a robust point completion method for its application in image-guided liver surgery.



\end{abstract}

\section{INTRODUCTION}
\label{sec:intro}  

To help surgeons locate critical surgical targets and structures, such as tumors and vessels, various registration methods have been developed \cite{heiselman2024image, yang2024boundary, yang2023learning}. Different registration approaches can be applied depending on the chosen intra-operative imaging modality. This study focuses on 3D-3D registration, where pre-operative data from CT or MRI, represented as a point cloud or mesh, is aligned with intra-operative data acquired through 3D sensors and reconstruction techniques \cite{yang2023disparity}, typically represented as a point cloud. However, this registration process is challenging due to partial intra-operative visibility of the liver, noise in the data, and liver deformation\cite{9414549,heiselman2018characterization}. Point cloud completion (PCC) methods aim to generate complete point clouds from partial point clouds, which can potentially improve registration accuracy by generating a cleaner and more complete intra-operative surface. Moreover, if these methods can directly recover the complete liver model from the intra-operative point cloud, they may even eliminate the need for traditional registration techniques. 



To our knowledge, only a few research studies have been conducted on liver intra-operative point cloud completion. 
Xi {\it et al.}\cite{xi2021recovering} primarily focused on recovering dense 3D point clouds from monocular depth estimation methods. However,  Xi {\it et al.} employed an affine transformation to deform the liver, which does not accurately capture realistic deformations. Whether point cloud completion methods can effectively handle realistic tissue deformations, such as those generated from biomechanical models, remains uncertain. In the framework proposed by Foti {\it et al.}{\it\cite{foti2020intraoperative}}, manual identification is still required to roughly select a region from the pre-operative model corresponding to the visible intra-operative surface. However, this manual identification process is not trivial and is prone to errors. Lastly, neither of the above two methods is publicly available, hence hindering the understanding of the performance of point cloud completion methods toward applications in image-guided liver surgery.

This study examines and evaluates several state-of-the-art point cloud completion methods for liver surgery applications under different cases. Specifically, we currently focus on the patient-specific approach to evaluate their performances, where simulated intra-operative point clouds are generated by deforming the single liver model. This simplification allows us to focus on other factors influencing completion results, such as visibility ratio and noise. In addition, a patient-specific approach could offer high accuracy.

We perform experiments that cover three main cases: canonical pose, non-canonical pose, and noisy data in the canonical pose. During surgical procedures, the camera field of view and patient positioning may lead to the intra-operative liver being captured from various poses. Point cloud completion methods are expected to generalize well to those arbitrary liver poses to be effective in the surgical environment. Thus, we evaluate the methods at canonical and non-canonical poses. The point clouds exhibit consistent positions and orientations in the canonical pose, whereas they have arbitrary positions and orientations in the non-canonical pose. To account for noisy intra-operative point cloud data from 3D sensors, we also simulated the noisy data condition by adding Gaussian noise to the canonical point clouds.



\section{Methodology}
\subsection{Problem Definition}

Given a partial intra-operative point cloud, we aim to find the optimal point cloud completion method to generate a complete point cloud accurately and robustly. The intra-operative liver point clouds obtained by deforming and cropping the pre-operative liver model in different ways are input to the six well-established learning-based point cloud completion methods: FoldingNet, PCN, GRNet, TopNet, PoinTr, and AdaPoinTr. 



\label{sec:sections}
 
\subsection{FoldingNet}
 FoldingNet\cite{yang2018foldingnet} is an auto-encoder network that performs folding operations on the decoder. The encoder consists of two graph layers that promote local structural information. The encoder's output is a high-dimensional embedding of an input point cloud called codewords. Folding operations start by augmenting 2D grid points with codewords. The surface structure is reproduced by combining multiple folding operations.
 
\subsection{PCN}
PCN\cite{yuan2018pcn} is also an encoder-decoder network. The encoder outputs feature vectors generated from a partial input point cloud. The decoder comprises a multi-stage point generation pipeline incorporating both fully connected and folding-based decoders. The former is responsible for generating a sparse point set that represents the global geometry of a shape. The latter is responsible for generating a more smooth surface that represents the local geometry of a shape. This combination enables PCN to effectively reconstruct an incomplete point cloud by balancing global structure and local detail.

\subsection{TopNet}
TopNet\cite{tchapmi2019topnet} introduced a novel rooted tree-structure decoder to model arbitrary structure/topology on a point cloud. The root node of the tree receives feature vectors as input from the encoder. Multi-layer perceptrons (MLPs) are arranged in a tree structure to produce feature vectors for their child nodes. The leaf nodes of the tree output the final 3D points.

\subsection{GRNet}
To fully capture structural and contextual information, GRNet\cite{xie2020grnet} introduces three key components: gridding, gridding reverse, and cubic feature sampling. The gridding layer creates 3D grids as intermediate representations to capture spatial relationships in the point cloud using 3D convolution. The gridding reverse layer generates the coarse point cloud from the 3D grid. Using feature maps from the 3D CNN layers, cubic feature sampling produces features for coarse point clouds. Finally, the MLP layer takes the coarse point cloud and the features to produce the final completed point cloud.

\subsection{PoinTr and AdaPoinTr }
PoinTr\cite{yu2021pointr} leverages a transformer encoder-decoder architecture to capture more detailed structural information. It converts the point cloud completion task into a set-to-set translation task, where point proxies of the partial point cloud are input and point proxies of the missing part are output. The point proxies are generated by grouping the point cloud into sets of points with positional embedding. It also introduces a geometry-aware block to model the local geometrical relationship. AdaPoinTr\cite{yu2023adapointrdiversepointcloud} is an extension of PoinTr with the addition of adaptive denoising queries.

\section{Experiments}
Experiments are conducted to evaluate the performance of six different PCC methods under three conditions: canonical pose (Sec.~\ref{sec:Canonical}), non-canonical pose (Sec.~\ref{sec:Non-Canonical}), and noisy data in the canonical pose (Sec.~\ref{sec:Noise}).

   \begin{figure} [ht]
   \begin{center}
   \begin{tabular}{c} 
   \includegraphics[width=0.9\textwidth]{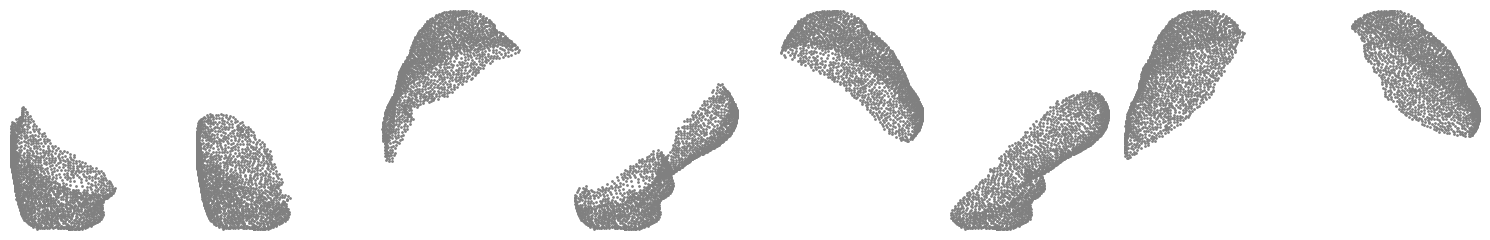}
   \end{tabular}
   \end{center}
   \caption[example] 
   { \label{fig:inputsVP}Illustration of eight different partial inputs generated for each liver model during testing. Eight distinct viewpoints are used to create these partial inputs, with missing points located in various regions.}
   \end{figure} 

\subsection{Intra-operative Liver Point Cloud Data}
The pre-operative liver point cloud model used in this study originates from the Phantom A model, as described by Yang {\it et al}.\cite{yang2024boundary}. To create intra-operative liver point clouds, we simulate deformations of the Phantom A model following the procedures outlined in our previous study\cite{yang2023learning}. The deformation fields are generated using a neo-Hookean hyperelastic material model, following the approach outlined by Pfeiffer {\it et al}\cite{pfeiffer2020non}. The deformations are simulated to generate deformed models using a finite element solver. In total, 5000 intra-operative liver models, each represented as a point cloud 
$P = \{ \mathbf{p}_i \mid \mathbf{p}_i = ({x}_i, {y}_i, {z}_i) \in \mathbb{R}^3 , i = 1, 2, \ldots, N \},$ consisting of $N = 8192$ points, are created by deforming the Phantom A model in different ways. 
\subsection{Point Cloud Normalization} 
To make the point cloud invariant to translation and scaling, each liver model is normalized based on its centroid and the maximum distance from the centroid. The centroid $\mathbf{c}$ of point cloud $P$ and the distance \( d_{\text{max}} \) from $\mathbf{c}$ to the point that is farthest from it is computed as Equation~(\ref{eq:c}).
\begin{equation}
\label{eq:c}
\mathbf{c} = \frac{1}{N} \sum_{i=1}^N \mathbf{p}_i, \quad d_{\text{max}} = \max_{i=1}^N \|\mathbf{p}_i - \mathbf{c}\|_2
\end{equation}
Finally, point cloud \( {P} \) is normalized by subtracting the centroid and dividing by \( d_{\text{max}} \) as given in Equation~(\ref{eq:pcnorm}).
\begin{equation}
\label{eq:pcnorm}
    {\tilde{P}} = \frac{{P} - \mathbf{c}}{d_{\text{max}}}
\end{equation}

\subsection{Experimental Setup}

The dataset is divided into training, validation, and testing sets in the ratio of 8:1:1. The partial point clouds are generated on the fly from the complete point clouds following the method proposed by Yu {\it et al.}\cite{yu2021pointr} for the ShapeNet-55 and ShapeNet-34 benchmarks. During the training and validation phases, a viewpoint is generated randomly, and a crop point is selected randomly between 2048 and 6144 (25\% to 75\% of the surface). During the testing phase, eight viewpoints are fixed to generate eight unique partial inputs per liver with missing points in different regions, as depicted in Figure~\ref{fig:inputsVP}. 

\begin{table}[ht]
\caption{Batch size and learning rate utilized for training different PCC methods.
} 
\setlength{\tabcolsep}{11pt}

\label{tab:Batch-LR}
\begin{center}       
\begin{tabular}{ccccccc}
\Xhline{1.2pt}
\rule[-1ex]{0pt}{3.5ex}   & FoldingNet & PCN & GRNet & TopNet & PoinTr & AdaPoinTr \\
\hline
\rule[-1ex]{0pt}{3.5ex}  Batch Size & 16 & 16 & 8 & 8 & 16 & 16 \\
\rule[-1ex]{0pt}{3.5ex}  Learning Rate & 0.0001 & 0.0001 & 0.0001 & 0.001 & 0.0005 & 0.0001 \\
\Xhline{1.2pt}
\end{tabular}
\end{center}
\end{table}

For canonical pose, the performance of the tested PCC methods is evaluated using partial inputs with varying missing points under three modes: simple (25\% of surface missing with 2048 points cropped), medium (50\% of surface missing with 4096 points cropped), and hard (75\% of surface missing with 6144 points cropped). This division aims to assess the models' performance in completing surfaces with varying amounts of missing regions. However, to isolate the impact of surface visibility on models performance, we restricted our experiments to only simple mode (25\% of surface missing with 2048 points cropped) for non-canonical pose and noisy data in canonical pose. The remaining points after cropping are down-sampled to 2048 points using furthest point sampling, which is the input data to the model. All the models have been trained for 150 epochs on NVIDIA A100 GPU with learning rate and batch size given in Table~\ref{tab:Batch-LR}. The codebase and other configurations utilized in this work are publicly available at  \href{https://github.com/yuxumin/PoinTr}{https://github.com/yuxumin/PoinTr}. The current settings make it easy for us to examine the performance of the PCC methods at this stage; however, these settings should be improved to address more realistic scenarios in future work.

\subsubsection{Non-canionical pose setup}


We generated non-canonical liver models by applying different randomly generated rotations between [0, 2$\pi$] to the canonical liver models. We utilized three train/test rotation settings: Z/Z, where both training and testing involve rotations along the Z-axis; Z/SO(3), where training involves rotations along the Z-axis, and testing involves arbitrary SO(3) rotations; and SO(3)/SO(3), where both training and testing involve arbitrary SO(3) rotations.


\subsubsection{Noise setup}

To introduce noise into the point cloud data, we apply a Gaussian perturbation to the original points constructing a noisy point cloud by sampling from a Gaussian distribution $N \sim \mathcal{N}(0, \sigma^2)$, where $\sigma \in \{0.5, 1, 3, 5 \}$ mm. The noise is added to the original points as: ${P}_{\text{noisy}} = {P} + {N}$.

\subsection{Evaluation Metrics}
Chamfer Distance (CD) is a commonly used metric in point cloud generation tasks that measures the bidirectional
similarity between a generated and a ground truth point cloud. As defined in Equation~(\ref{eq:cd}), CD computes the average closest distance from each point in the prediction to the ground truth, and vice versa,
\begin{equation}
\label{eq:cd}
d_{CD}(\mathcal{P}, \mathcal{G}) = \frac{1}{|\mathcal{P}|} \sum_{p \in \mathcal{P}} \min_{g \in \mathcal{G}} ||p - g||_2 + \frac{1}{|\mathcal{G}|} \sum_{g \in G} \min_{p \in P} ||g - p||_2
\end{equation}
where $|\mathcal{P}|$ and $|\mathcal{G}|$ represent the total number of points in the prediction and ground truth sets, respectively, and $d_{CD}(\mathcal{P}, \mathcal{G})$ denotes the Chamfer Distance between the point sets $\mathcal{P}$ and $\mathcal{G}$.

Hausdorff Distance (HD) measures the largest discrepancy between prediction and ground truth point clouds. It is computed by determining the maximum nearest distance from each point in the prediction to the ground truth and vice versa, and again taking the maximum value between these two distances,

\begin{equation}
\label{eq:hd}
d_{HD}(\mathcal{P}, \mathcal{G}) = \max \left\{ 
\max_{p \in \mathcal{P}} \left\{ \min_{g \in \mathcal{G}} \| p - g \|_2 \right\},\; 
\max_{g \in \mathcal{G}} \left\{ \min_{p \in \mathcal{P}} \| g - p \|_2 \right\}
\right\}
\end{equation}
where $d_{HD}(\mathcal{P}, \mathcal{G})$ denotes the Hausdorff Distance between the point sets $\mathcal{P}$ and $\mathcal{G}$.

The F-Score at a 1 mm threshold $(d)$ is utilized as another evaluation metric. It is calculated as the harmonic mean of precision and recall. Precision measures the percentage of predicted points that lie within 1 mm of the nearest ground truth points. Similarly, recall measures the percentage of ground truth points that lie within 1 mm of the nearest predicted points,
\begin{equation}
\text{F-Score}(d) = \frac{2P(d)R(d)}{P(d) + R(d)}
\end{equation}
where $P(d)$ and $R(d)$ denote the precision and recall.


\section{Results}

\subsection{Results on Canonical Pose} 
\label{sec:Canonical}

\begin{figure} [th]
\begin{center}
\begin{tabular}{c} 
\includegraphics[width=0.95\textwidth]{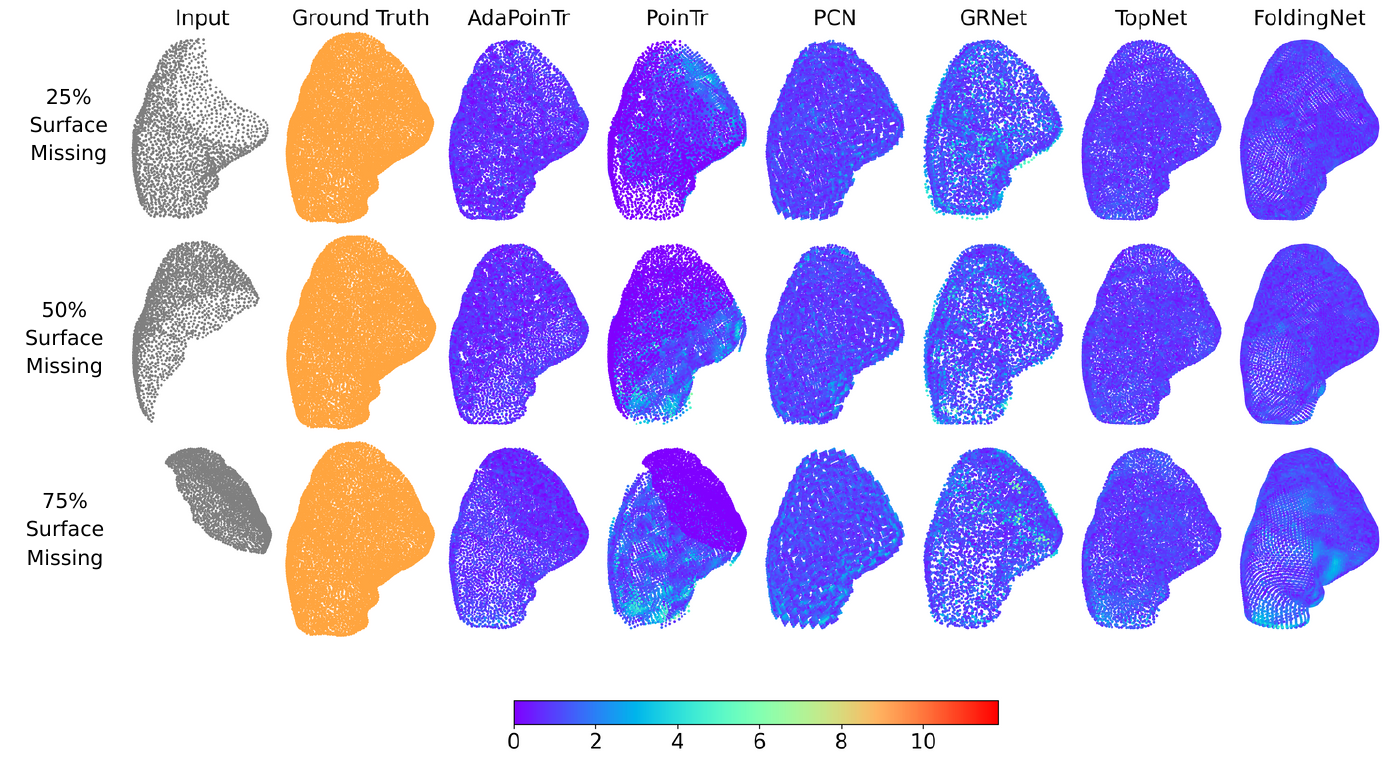}   
\end{tabular}
\end{center}
\caption[example] 
{ \label{fig:output} 
Qualitative comparison of point cloud completion results on canonical liver models. The first column shows the input incomplete point clouds, with 25\% missing regions in the first row, 50\% missing regions in the second, and 75\% missing regions in the third. The second column shows the ground truth complete point clouds. The subsequent columns present the completion results from various PCC methods: AdaPoinTr, PoinTr, PCN, GRNet, TopNet, and FoldingNet. The colorbar represents the Euclidean Distance (in mm) from each point in the prediction to its nearest point in the ground truth.
}
\end{figure} 

\begin{table}[ht]
\caption{Performance of PCC methods on canonical liver models. CD-S, CD-M, and CD-H are Chamfer Distances (in mm) for the simple, medium, and hard modes when the partial input liver consists of 75\%, 50\%, and 25\% of the total surface. Similarly, HD-S, HD-M, and HD-H are the Hausdorff Distance (in mm) for simple, medium, and hard modes. F-Score is the average value of three modes at a 1 mm threshold. *p $< 0.05$ indicates a statistically significant improvement in terms of the performance yielded by a particular method over the other methods.
} 
\label{tab:opstd}

\begin{center}  
\small

\begin{tabular}{cccccccc}

\Xhline{1.2pt}
\rule[-1ex]{0pt}{3.5ex}  & CD-S & CD-M & CD-H & HD-S & HD-M & HD-H & F-Score \\ 
\hline

\rule[-1ex]{0pt}{3.5ex} FoldingNet & $2.30 \pm 0.52$ & $2.48 \pm 0.63$ & $2.81 \pm 0.82$ & $6.34 \pm 2.60$ & $7.00 \pm 3.11$ & $8.04 \pm 3.91$ & $0.43$ \\

\rule[-1ex]{0pt}{3.5ex} TopNet & $2.22 \pm 0.50$ & $2.33 \pm 0.58$ & $2.58 \pm 0.76$ & $5.51 \pm 2.94$ & $6.82 \pm 3.85$ & $8.50 \pm 4.32$ & $0.47$ \\

\rule[-1ex]{0pt}{3.5ex} GRNet & $2.82 \pm 0.24$ & $2.84 \pm 0.34$ & $2.99 \pm 0.51$ & $7.36 \pm 1.83$ & $7.92 \pm 2.67$ & $8.89 \pm 3.70$ & $0.34$ \\

\rule[-1ex]{0pt}{3.5ex} PCN & $2.47 \pm 0.44$ & $2.64 \pm 0.52$ & $2.92 \pm 0.69$ & $5.83 \pm 2.51$ & $6.52 \pm 2.99$ & $7.49 \pm 3.70$ & $0.39$ \\

\rule[-1ex]{0pt}{3.5ex} PoinTr & $2.34 \pm 0.47$ & $2.56 \pm 0.60$ & $2.92 \pm 0.75$ & $8.10 \pm 3.02$ & $8.84 \pm 3.47$ & $10.09 \pm 3.78$ & $0.43$ \\

\rule[-1ex]{0pt}{3.5ex} AdaPoinTr & $^ * 1.62 \pm 0.15$ & $^ * 1.73 \pm 0.26$ & $^ * 2.00 \pm 0.45$ & $^ * 4.22 \pm 1.86$ & $^ * 5.14 \pm 3.05$ & $^ * 7.27 \pm 4.17$ & $^ * 0.63$ \\

\Xhline{1.2pt}
\end{tabular}
\end{center}
\end{table}

\begin{table}[ht!]
\caption{Performance of PCC methods on non-canonical liver models. Z/Z denotes training and testing with rotations along the Z-axis, Z/SO(3) involves training with Z-axis rotations and testing with arbitrary 3D rotations, and SO(3)/SO(3) represents both training and testing with arbitrary 3D rotations. CD and HD represents the Chamfer Distance (in mm) and Hausdorff Distance (in mm), respectively. *p $< 0.05$ indicates a statistically significant improvement in
terms of the performance yielded by a particular method over the other methods.}

\label{tab:Rotation}
\begin{center}       
\setlength{\tabcolsep}{18pt}

\begin{tabular}{c c c c c c} 
\Xhline{1.2pt}
       
            \rule[-1ex]{0pt}{3.5ex}
            &  & Z/Z & Z/SO(3) & SO(3)/SO(3) \\

        \hline
        \rule[-1ex]{0pt}{3.5ex}
        \multirow{3}{*}{FoldingNet} & CD &  $2.86 \pm 0.59$ & $13.24 \pm 4.24$ & $4.60 \pm 0.59$ \\
                                    
                                    & HD & $8.40 \pm 2.74$ & $28.60 \pm 9.60$ & $13.33 \pm 2.38$ \\
                                    & F-Score & $0.35 \pm 0.08$ & $0.07 \pm 0.04$ & $0.14 \pm 0.03$ \\

        \hline
        
        \rule[-1ex]{0pt}{3.5ex}
        \multirow{3}{*}{TopNet}     & CD & $2.71 \pm 0.56$ & $11.71 \pm 4.10$ & $4.89 \pm 0.60$ \\
                                    & HD & $7.51 \pm 2.58$ & $26.49 \pm 7.52$ & $14.99 \pm 2.44$ \\
                                    & F-Score & $0.36 \pm 0.09$ & $0.04 \pm 0.04$ & $0.12 \pm 0.03$ \\
        \hline
        \rule[-1ex]{0pt}{3.5ex}
        \multirow{3}{*}{GRNet}      & CD  & $2.98 \pm 0.20$ & $4.59 \pm 0.69$ & $3.14 \pm 0.17$ \\
                                    & HD & $7.65 \pm 1.87$ & $21.23 \pm 5.21$ & $8.96 \pm 2.19$ \\
                                    & F-Score & $0.31 \pm 0.03$ & $0.23 \pm 0.02$ & $0.29 \pm 0.02$ \\

        \hline
        \rule[-1ex]{0pt}{3.5ex}
        \multirow{3}{*}{PCN}        & CD & $3.00 \pm 0.47$ & $11.28 \pm 3.25$ & $4.39 \pm 0.57$ \\
                                    & HD & $7.09 \pm 2.50$ & $24.61 \pm 6.33$ & $11.38 \pm 2.39$ \\
                                    & F-Score & $0.31 \pm 0.06$ & $0.07 \pm 0.04$ & $0.15 \pm 0.03$ \\

        \hline
        \rule[-1ex]{0pt}{3.5ex}
        \multirow{3}{*}{PoinTr}     & CD & $2.40 \pm 0.36$ & $7.14 \pm 1.64$ & $2.12 \pm 0.23$ \\
                                    
                                    & HD & $8.52 \pm 2.65$ & $25.99 \pm 5.94$ & $7.99 \pm 2.29$ \\
                                    & F-Score& $0.43 \pm 0.04$ & $0.29 \pm 0.02$ & $0.47 \pm 0.03$ \\
        \hline
        \rule[-1ex]{0pt}{3.5ex}
        \multirow{3}{*}{AdaPoinTr}  & CD & $^ * 1.75 \pm 0.15$ & $^ * 4.17 \pm 1.28$ & $^ * 1.88 \pm 0.15$ \\
                                    
                                    & HD & $^ *5.23 \pm 1.92$ & $^ *19.16 \pm 6.49$ & $^ *5.93 \pm 2.06$ \\
                                    & F-Score & $^ *0.63 \pm 0.05$ & $^ *0.36 \pm 0.08$ & $^ *0.59 \pm 0.04$ \\
\Xhline{1.2pt}

\end{tabular}
\end{center}
\end{table}

\begin{table}[ht!]
\caption{Performance of PCC methods on canonical liver models with noise. Gaussian noise is applied with mean $0$ and standard deviations of $0.5$ mm, $1$ mm, $3$ mm, and $5$ mm. CD and HD represent the Chamfer Distance (in mm) and Hausdorff Distance (in mm), respectively. *p $< 0.05$ indicates a statistically significant improvement in
terms of the performance yielded by a particular method over the other methods.}
\setlength{\tabcolsep}{12pt}

\label{tab:noise}

\begin{center}       
\begin{tabular}{cccccc} 

\Xhline{1.2pt}
        \rule[-1ex]{0pt}{3.5ex}
         &  &$0.5$& $1$ & $3$ & $5$ \\
        \hline
        \rule[-1ex]{0pt}{3.5ex}
        \multirow{3}{*}{FoldingNet} & CD & $2.31 \pm 0.51$ & $2.62 \pm 0.47$ & $6.03 \pm 0.51$ & $10.19 \pm 0.83$ \\
        
        & HD & $6.42 \pm 2.52$ & $6.63 \pm 2.44$ & $8.80 \pm 2.32$ & $13.77 \pm 2.10$ \\
        & F-Score  & $0.48 \pm 0.12$ & $0.36 \pm 0.07$ & $0.08 \pm 0.02$ & $0.04 \pm 0.01$ \\

        \hline
        \rule[-1ex]{0pt}{3.5ex}
        \multirow{3}{*}{TopNet}     & CD & $2.34 \pm 0.48$ & $2.80 \pm 0.43$ & $6.02 \pm 0.47$ & $8.58 \pm 0.77$  \\
        & HD& $5.56 \pm 2.92$ & $6.19 \pm 2.91$ & $12.50 \pm 2.89$ & $17.85 \pm 3.28$ \\
        & F-Score  & $0.44 \pm 0.11$ & $0.29 \pm 0.05$ & $0.09 \pm 0.02$ & $0.06 \pm 0.01$ \\
        \hline
        \rule[-1ex]{0pt}{3.5ex}
        \multirow{3}{*}{GRNet}      & CD& $2.89 \pm 0.24$ & $3.11 \pm 0.23$ & $5.62 \pm 0.23$ & $7.99 \pm 0.39$  \\
                                   
                                    & HD& $7.52 \pm 1.73$ & $8.13 \pm 1.57$ & $13.87 \pm 2.31$ & $19.20 \pm 2.04$ \\
                                     & F-Score & $0.32 \pm 0.05$ & $0.28 \pm 0.04$ & $0.10 \pm 0.01$ & $0.06 \pm 0.01$ \\

        \hline
        \rule[-1ex]{0pt}{3.5ex}
        \multirow{3}{*}{PCN}        & CD & $2.53 \pm 0.43$ & $2.83 \pm 0.41$ & $5.71 \pm 0.46$ & $10.15 \pm 0.65$ \\
                                    
                                    & HD& $5.92 \pm 2.50$ & $6.23 \pm 2.44$ & $10.19 \pm 2.07$ & $16.27 \pm 2.02$ \\
                                    & F-Score  & $0.40 \pm 0.08$ & $0.31 \pm 0.06$ & $0.09 \pm 0.02$ & $0.04 \pm 0.01$ \\
        \hline
        \rule[-1ex]{0pt}{3.5ex}
        \multirow{3}{*}{PoinTr}     & CD & $2.75 \pm 0.46$ & $3.19 \pm 0.47$ & $^ * 4.98 \pm 0.61$ & $ ^ * 6.60 \pm 0.77$  \\
                                  
                                    & HD & $8.13 \pm 2.96$ & $8.30 \pm 2.88$ & $^ * 12.18 \pm 1.86$ & $^ * 19.17 \pm 1.60$  \\
                                      & F-Score& $0.39 \pm 0.05$ & $0.26 \pm 0.05$ & $ ^ * 0.12 \pm 0.03$ & $ ^ * 0.08 \pm 0.02$ \\
        \hline
        \rule[-1ex]{0pt}{3.5ex}
        \multirow{3}{*}{AdaPoinTr}  & CD & $^ *1.91 \pm 0.15 $ & $2.52 \pm 0.16$ & $6.57 \pm 0.50$ & $9.05 \pm 0.85$  \\
                                    
                                    & HD & $^ *4.31 \pm 1.79$ & $^ * 5.10 \pm 1.92$ & $12.55 \pm 1.63$ & $19.83 \pm 1.70$ \\
                                    & F-Score & $^ *0.59 \pm 0.05$ & $ ^ * 0.35 \pm 0.04$ & $0.06 \pm 0.01$ & $0.04 \pm 0.01$ \\
\Xhline{1.2pt}
\end{tabular}
\end{center}
\end{table}

 Table~\ref{tab:opstd} shows the Chamfer Distance, Hausdorff Distance, and F-Score values at the 1 mm threshold. CD-S, CD-M, and CD-H denote the Chamfer Distance corresponding to the simple, medium, and hard modes, respectively, with partial liver input of 75\%, 50\%, and 25\% of the total surface. Similarly, HD-S, HD-M, and HD-H denote the Hausdorff Distance for simple, medium, and hard modes. F-Score is the average of three modes. 
 
 Among all tested PCC methods, PCN and GRNet yielded high CD values and low F-Scores, demonstrating poor performance. PoinTr and FoldingNet performed
slightly better than these methods with moderate CD and F-Score values. AdaPoinTr outperformed all other methods, achieving superior results across all metrics. Except for PCN and PoinTr, the performance trend across all methods, evaluated using HD, CD, and F-Score, are nearly identical. Despite its poor performance in terms of CD and F-Score, PCN achieved low HD values, particularly in the medium and hard modes. In contrast, PoinTr exhibited significantly higher HD values. The increase in CD and HD values from the simple mode (75\% intra-operative liver surface visibility) to hard mode (25\% intra-operative liver surface visibility) across all models indicates that reduced surface coverage in the input lowers the completion performance of PCC methods, which is not unexpected.

\

 Figure~\ref{fig:output} presents input with varying missing surface, ground truth, and output from different models. The points in the prediction are color-coded based on their distance to the nearest point in the ground truth, with the least distance represented by violet. AdaPoinTr exhibits many points that perfectly align with the ground truth. This observation also aligns with the lower CD and higher F-Score values for AdaPoinTr. Additionally, the visualization clearly shows that the models struggle to generate points in regions with missing points. For regions where points are present, the prediction closely matches the ground truth. This also suggests that a smaller input surface corresponds to a decrease in model performance, as observed above, with the increase in CD and HD values from the simple mode to the hard mode across all models.

\subsection{Results on Non-Canonical Pose} 
\label{sec:Non-Canonical}

The performance of PCC methods under non-canonical pose is identified by setting up three rotational setups as presented in Table~\ref{tab:Rotation}. Among them, AdaPoinTr outperformed other methods, showcasing robust performance to rotation in all modes with lower CD, HD values, and higher F-Score. Compared to the results of the Z/Z setting, their performance exhibits significant degradation at the Z/SO(3). In the SO(3)/SO(3), incorporating data augmentation during training improved performance across models. However, it is infeasible to include all possible SO(3) inputs during data augmentation, and models may fail when encountering unseen rotations during inference. 



\subsection{Results on Canonical Pose with Noise} 
\label{sec:Noise}

The performance of PCC methods under noisy data conditions is shown in Table~\ref{tab:noise}. CD and HD denote the Chamfer Distance and Hausdorff Distance, respectively. Gaussian noise with a mean of 0 and standard deviations of 0.5 mm, 1 mm, 3 mm, and 5 mm is added to the input data. At a noise level of 0.5 mm, FoldingNet demonstrated performance similar to its noise-free results (CD-S in Table~\ref{tab:opstd}). With increasing noise levels, PoinTr exhibited a gradual rise in CD values. In contrast, AdaPoinTr showed a substantial increase in CD values, reaching $6.57$ mm at a noise level of $3$ mm, where PoinTr outperformed it across all the metrics. At higher noise levels ($\ge$ 3), PoinTr showcased robust performance among all PCC methods tested.

\section{Conclusion and Future work}

We analyzed the point cloud completion capability of six deep learning-based models under three clinically-relevant scenarios. Our results on the canonical and non-canonical pose render AdaPoinTr as the most promising method. However, PoinTr is best able to handle higher noise levels. The performance of PCC methods deteriorated in the presence of arbitrary SO(3) rotations and high noise levels, which remain concerns for their applications in image-guided surgery. 

Future work will focus on verifying their performances in more realistic settings, designing rotation-equivariant architectures, and developing noise-resilient algorithms to address these challenges. Additionally, the integration of PCC methods into the registration process will be explored.






\acknowledgments 
 We would like to acknowledge the generous support for this work by the National Institutes of Health – National Institute of General Medical Sciences under Award No. R35GM128877 and the National Science Foundation - Division of Chemical, Bioengineering and Transport Systems under Award No. 2245152.


\bibliography{report} 
\bibliographystyle{spiebib} 

\end{document}